\def\year{2019}\relax
\documentclass[letterpaper]{article} 
\usepackage{aaai19}  
\usepackage{times}  
\usepackage{helvet}  
\usepackage{courier}  
\usepackage{url}  
\usepackage{graphicx}  
\usepackage{latexsym}
\usepackage{amsfonts}
\usepackage{amsmath}
\usepackage{amssymb}
\usepackage{latexsym}
\usepackage{booktabs}
\usepackage{array}
\usepackage{subcaption}
\usepackage[export]{adjustbox}
\usepackage[font=small,labelfont=bf]{caption}

\newcolumntype{L}{>{\centering\arraybackslash}m{4.5cm}}

\DeclareMathOperator*{\argmin}{argmin}

\begin{document}
%
\title{GlobalTrait: Personality Alignment of Multilingual Word Embeddings}
\author{Farhad Bin Siddique $^{12}$, Dario Bertero $^{12}$, Pascale Fung $^{123}$\\
$^1$ Electronic and Computer Engineering Department\\
$^2$ Center for Artificial Intelligence Research (CAiRE)\\
$^3$ EMOS Technologies Inc.\\
The Hong Kong University of Science and Technology\\
Clear Water Bay, Hong Kong \\
\tt{[fsiddique, dbertero]@connect.ust.hk, pascale@ece.ust.hk}
}

\maketitle
\begin{abstract}
We propose a multilingual model to recognize Big Five Personality traits from text data in four different languages: English, Spanish, Dutch and Italian. Our analysis shows that words having a similar semantic meaning in different languages do not necessarily correspond to the same personality traits. Therefore, we propose a personality alignment method, GlobalTrait, which has a mapping for each trait from the source language to the target language (English), such that words that correlate positively to each trait are close together in the multilingual vector space. Using these aligned embeddings for training, we can transfer personality related training features from high-resource languages such as English to other low-resource languages, and get better multilingual results, when compared to using simple monolingual and unaligned multilingual embeddings. We achieve an average F-score increase (across all three languages except English) from 65 to 73.4 (+8.4), when comparing our monolingual model to multilingual using CNN with personality aligned embeddings. We also show relatively good performance in the regression tasks, and better classification results when evaluating our model on a separate Chinese dataset.
\end{abstract}

\section{Introduction}

According to \cite{allport1937personality}, personality refers to the characteristic pattern in a person's thinking, feeling, and decision making. It is a quality of a person across a relatively long period and is different from emotions, which can be perceived in the moment. We can think of it as, personality is to emotion what climate is to weather. The Big Five model of personality \cite{goldberg:93} is a common way of quantifying a person's personality, and is recognized by most psychologists around the world. It tries to represent the traits as scores across five dimensions: 

\begin{itemize}
    \itemsep0.1em
    \item \textbf{Extraversion vs Introverted} (\textit{Extr}) - sociable, assertive, playful vs aloof, reserved, shy;
    \item \textbf{Conscientiousness vs Unconscientious} (\textit{Cons}) - self-disciplined, organised vs inefficient, careless;
    \item \textbf{Agreeableness vs Disagreeable} (\textit{Agr}) - friendly, cooperative vs antagonistic, faultfinding;
    \item \textbf{Neuroticism vs Emotionally Stable} (\textit{Emot}) - insecure, anxious vs calm, unemotional;
    \item \textbf{Openness to Experience vs Cautious} (\textit{Openn}) - intellectual, insightful vs shallow, unimaginative.
\end{itemize}

Personality traits affect the usage of language in people \cite{mairesse2007using}, and it is an integral part of human-human interaction \cite{long2000personality,berry2000affect}. As we develop smarter dialogue systems, future virtual agents need to detect and adapt to different user personalities in order to express empathy \cite{fung2016towards}. Although traditionally the user personality can
be identified by having the user fill out a self-assessment
form such as the NEO Personality Inventory \cite{costa:2008}, this method is not feasible for many applications where we
may wish to identify user personality, such as dialogue
systems. Therefore, work on automatic personality assessment has become increasingly important recently with the rise in popularity of applications such as Human Resources (HR) screening, personalized marketing, and other social media related
user-profiling. 

\begin{figure*}
\centering
\begin{subfigure}{.5\textwidth}
\centering
  \includegraphics[scale=0.225]{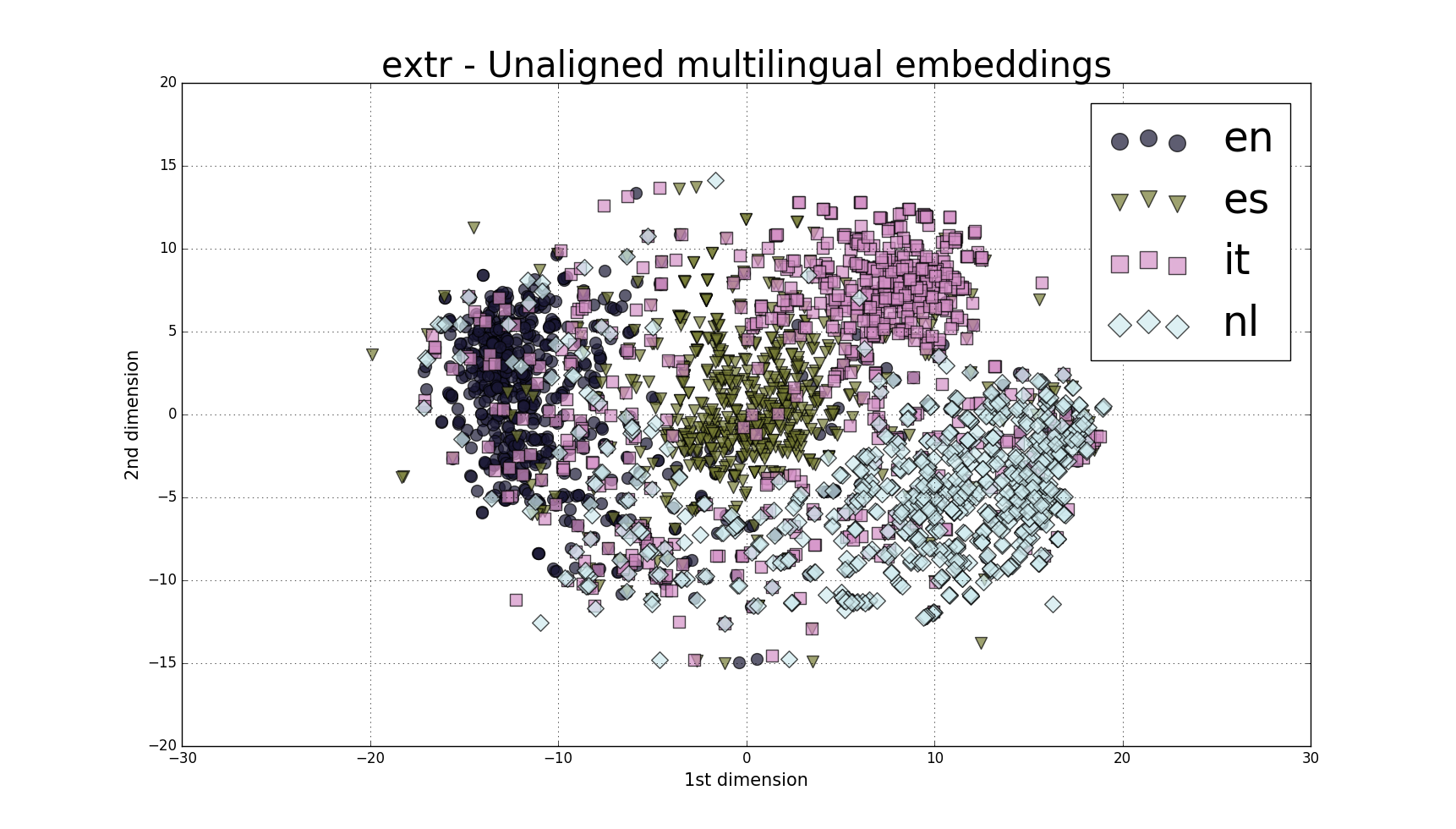}
  \caption{Words corresponding positively to Extraversion trait}
  \label{fig:sub1}
\end{subfigure}%
\begin{subfigure}{.5\textwidth}
\centering
  \includegraphics[scale=0.22]{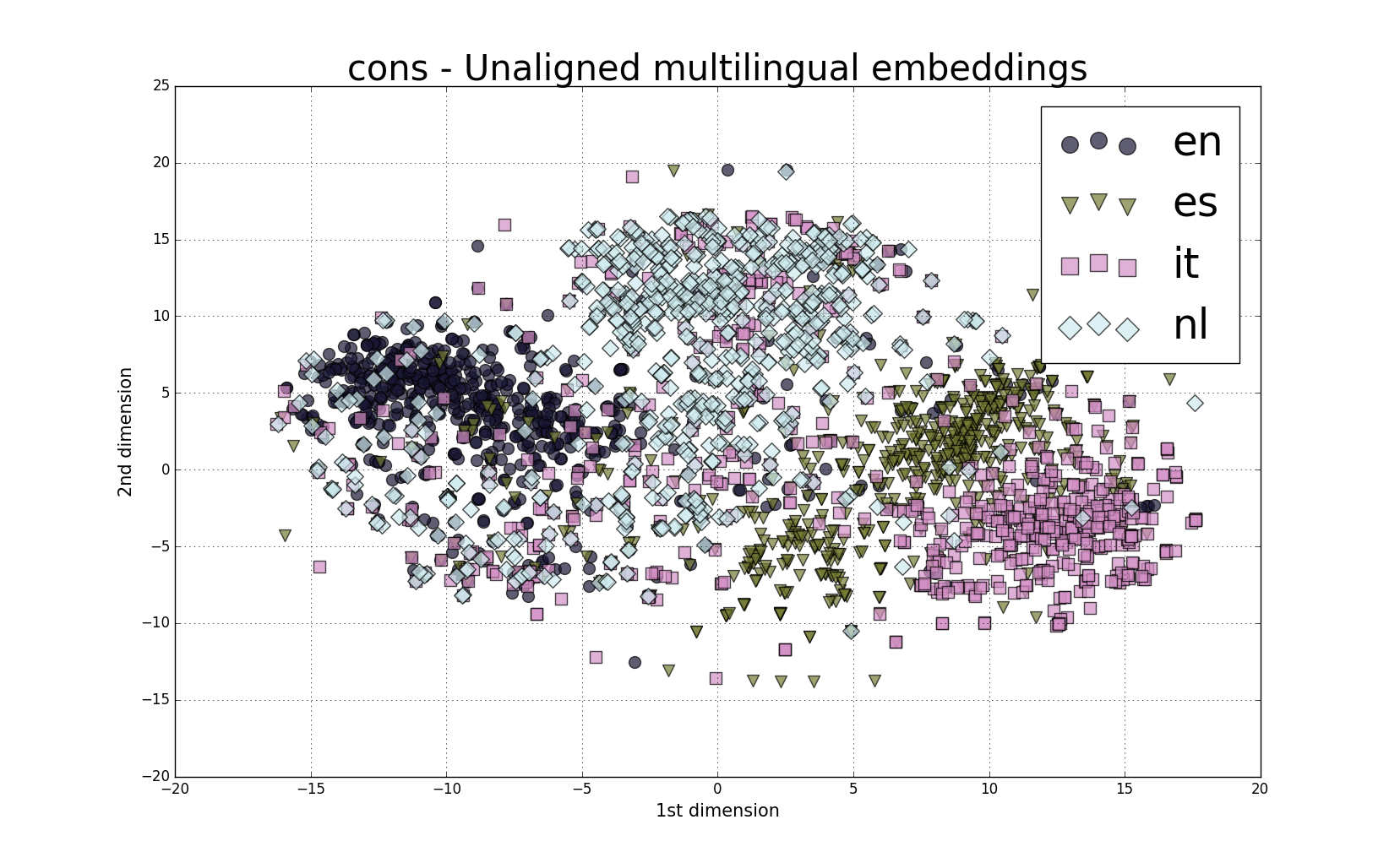}
  \caption{Words corresponding positively to Conscientiousness trait}
  \label{fig:sub2}
\end{subfigure}
\caption{Multilingual embeddings of most significant words corresponding to the Extraversion and Conscientiousness trait for the four languages, plotted in 2-Dimension using t-SNE. We see that positively correlated words for each trait are clustered per language, which shows that words corresponding to each language has different semantic/contextual meaning compared to the other languages. Therefore, we learn a mapping from each source language to our target language English, so the words are aligned based on each of the personality traits.}
\label{extr}
\end{figure*}

Currently, personality labeled data is scarce, especially in the multilingual setting, which makes it essential to use the relatively larger size of data in English to help recognize traits in other languages. Previous work on multilingual personality recognition have tried to use the word-level or character-level similarity across languages to develop a multilingual model \cite{liu2016language,siddique2017bilingual}. However, our experiments show that words used by people with different personality traits differ among languages or cultures, which is not captured by distributional semantics alone, making it necessary to learn a personality-based mapping of words to express each personality trait. Therefore, we propose GlobalTrait, which trains personality trait-based alignment of multilingual embeddings from the source language(s) to the target language, such that words that correlate positively to each trait are closer together in the multilingual (global) vector space. We show that taking such mapping or alignment of embeddings as input to our model gives us better multilingual results in the task of personality recognition.

\section{Related Work}

Automatic personality recognition has been done since as early as 2006 \cite{oberlander2006whose}, where the personality of blog authors were identified using Naive Bayes algorithm
and n-gram features. \cite{mairesse2007using} used two sources of lexical features, Linguistic Inquiry and Word Count (LIWC) \cite{pennebaker2001linguistic} and Medical Research Council (MRC) Psycholinguistics Database \cite{coltheart1981mrc} features, to identify personality from written and spoken transcripts. More recently, for tasks such as the Workshop on Computational Personality Recognition \cite{celli2013workshop}, people have worked on identifying personality from social media
texts (Facebook status updates and Youtube vlog transcriptions). \cite{verhoeven2013ensemble} used 2000 frequent trigrams as features and trained a SVM classifier, and \cite{farnadi2013recognising} used LIWC features to train SVM, Naive Bayes and K Nearest Neighbor (KNN) algorithms.

Deep Learning models such as Convolutional
Neural Networks (CNNs) have gained popularity
in the task of text classification \cite{kalchbrenner2014convolutional,kim2014convolutional}. This is because CNNs are good at capturing text features via its convolution operation, which can be applied on the text by taking the distributed representation of the words, called word embeddings, as input. Learning such distributed representation comes from the hypothesis that words that appear in similar contexts have similar meaning \cite{harris1954distributional}. Different works have been carried out in the past to learn such representations of words, such as \cite{mikolov2013distributed,pennington2014glove}, and more recently \cite{bojanowski2016enriching}. Cross-lingual or multilingual word embeddings try to capture such semantic information of words across two or more languages, such
that the words that have similar meaning in different languages are close together in the vector space \cite{faruqui2014improving,upadhyay2016cross}. For our task we use a more recent approach \cite{conneau2017word}, which does not require parallel data and learns a mapping from the source language embedding space to the target language in an unsupervised fashion.

\section{Methodology}

\begin{table*}[t]
  \caption{Some examples of significant words corresponding positively to each trait per language.}
  \label{words}
  \centering
   \scalebox{0.8}{
  \begin{tabular}{|l|L|L|L|L|}
  \toprule
         & English (En) & Spanish (Es) & Italian (It) & Dutch (Nl) \\
         \hline
        Extr & people, music, stories, followers, social & enhorabuena, gente, sociales, programa, tenemos & piaciuto, notizia, pubblicato, università, generazione & mensen, kennen, schoten, benieuwd, helemaal \\
        \hline
        Cons & research, work, time, analysis, project & gobierno, interesante, cuenta, contra, tiempo & direttori, giornalismo, palestra, rivoluzione, ricerca & raadscommissie, bibliotheek, grootdebat, harlopen, ambtenaren\\
        \hline
        Agr & life, feel, thanks, amazing, heart, right & gracias, estoy, estudiar, viernes, dormir & anche, siamo, utilizzato, fatto, grazie & inderdaad, altijd, genieten, hopelijk, geslaagd\\
        \hline
        Emot & want, today, really, shit, know, cause & pero, porque, alguien, sólo, vamos, quieres & sempre, proprio, fumetto, altro, scoprire & alleen, vragen, eigenlijk, waarom, volgande\\
        \hline
        Openn & love, new, watching, youth, skills, travel & mundo, auditori, trabajo, corazon, nuevo & mostra, mondo, aiutarmi, informati, completando & andere, iedereen, leven, lekker, misschien, volgende\\
    \bottomrule
  
  \end{tabular}
  }
\end{table*}

We propose GlobalTrait, a text-based model that uses multilingual embeddings across languages to train personality alignment per trait, such that words in the languages that correspond positively to one trait are closer together in the multilingual vector space: which are then fed to a CNN model for binary classification of each trait. We first train multilingual embeddings in the given languages, followed by identifying the most significant words in each language corresponding positively to each trait, which are used to learn a mapping from each source to the target language. The target language is the language in which we have the most labeled data available, in our case English. The mapping is essentially a personality trait based alignment that tries to bring the distributional representation of words correlating positively to each trait closer together. The initial multilingual embeddings along with the GlobalTrait aligned embeddings are then fed into a two channel CNN model to extract the relevant features for classification via a fully connected layer followed by softmax.

\subsection{Multilingual Embeddings Training}

Learning distributed representation of words (word embeddings) comes from the hypothesis that words that appear in similar contexts have similar meaning \cite{harris1954distributional}. Different works have been done to learn such representations \cite{mikolov2013distributed,pennington2014glove,bojanowski2016enriching}. Cross-lingual or multilingual word embeddings try to capture such semantic information of words across two or more languages \cite{faruqui2014improving,upadhyay2016cross}. We use the methodology of Multilingual Unsupervised and Supervised Embeddings (MUSE) \cite{conneau2017word} to first train multilingual embeddings across the four languages. MUSE tries to learn a mapping $W$ of dimension $d \times d$, where $d=300$ is the embedding dimension we use, such that:

\begin{equation}
W^* = \argmin_{W \in O_d(\mathbb{R})} || WX - Y ||_F
\end{equation}

where, $O_d(\mathbb{R})$ suggests that $W$ is an orthogonal matrix consisting of real numbers, $X$ and $Y$ are the $d \times n$ matrices representing the word embeddings of $n$ words in the source and target languages respectively. It is important that the mapping matrix $W$ is orthogonal, so that we are performing a rotational mapping on the embedding space, which does not disrupt the monolingual semantic information of the original embeddings. Also, being orthogonal gives us the following Procrustes solution to equation (1):

\begin{equation}
W^* = UV^T, \text{with } U\Sigma V^T = SVD(YX^T). 
\end{equation}

$W$ mapping matrix is trained via a Generative Adversarial Network (GAN) training approach \cite{goodfellow2014generative,ganin2016domain}, where a generator and a discriminator network are both trained in parallel. It is a two player game where the discriminator tries to differentiate between a source embedding and a mapped embedding, and the generator tries to fool the discriminator by making $WX$ as similar to $Y$ as possible. For our monolingual embeddings used for each of the individual languages, we use the pre-trained word embeddings via fastText\footnote{\tt https://github.com/facebookresearch/\\
fastText}. The issue with multilingual embeddings is that it only captures the semantic information of words across languages, so words that share similar context will appear close together in the multilingual vector space. As we can see in Figure \ref{extr} (explained further in the \textit{Experiments} section), words that correspond to the same trait do not always share a similar semantic meaning across languages. This motivates the need for a mapping of the embeddings from the source to the target language, hence the notion of personality trait-based alignment.

\begin{figure}[t]
\centering
\includegraphics[scale=0.4]{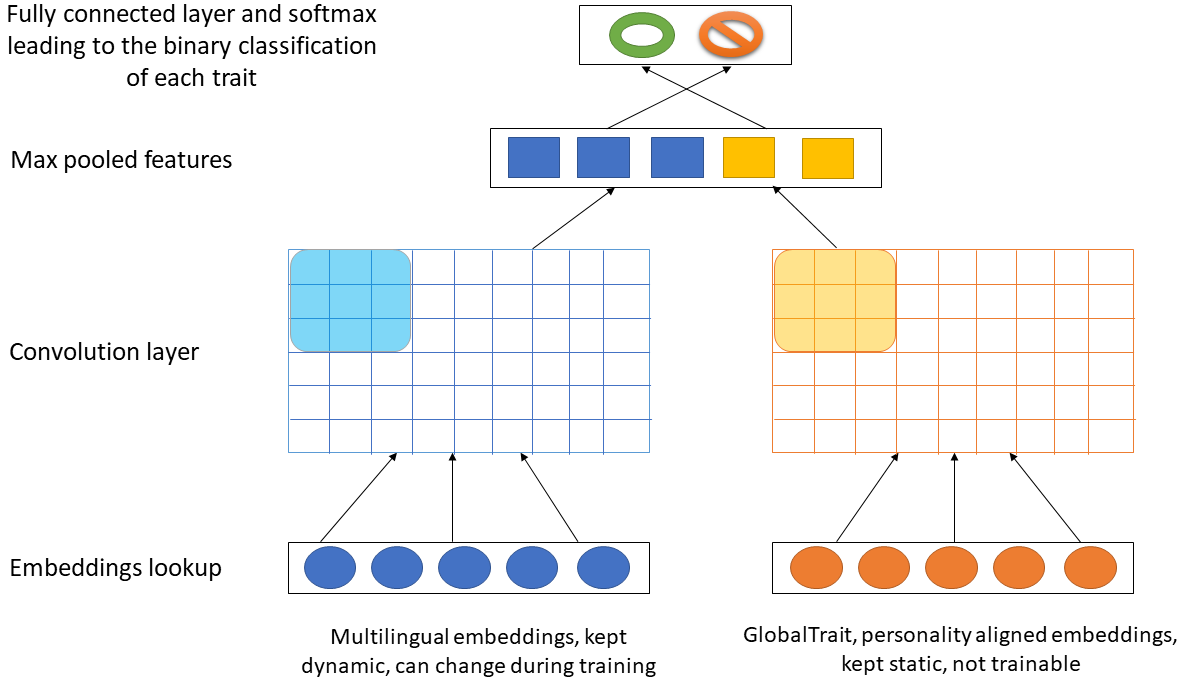}
\caption{Simplified architecture of our two channel CNN model, one channel taking the unaligned multilingual embeddings as input, while the other taking the GlobalTrait aligned embeddings, fed into a one layer CNN, and the extracted features are concatenated, followed by a max-pooling layer and a fully connected layer to softmax for binary classification.}
\label{cnn}
\end{figure}

\subsection{GlobalTrait - One Alignment Per Trait}

We propose GlobalTrait, which does a personality trait-based alignment of the multilingual embeddings from the source to the target language. From the dataset, we use Term Frequency - Inverse Document Frequency (\textit{tf-idf}) features to obtain the \textit{n} most significant words that correspond positively to each trait per language, and get the multilingual embeddings corresponding to the words. Using these embeddings, we learn a second mapping from each source language to our target language, English, with the idea that the mapped or aligned embeddings will represent the trait, by being closer together in the vector space. There has to be one alignment per trait per language - a rotational mapping from the source to the target language space. We end up training 5 different mapping matrices for each source language, one for each of the Big Five traits.

\begin{figure*}[t]
\centering
\includegraphics[scale=0.29,left]{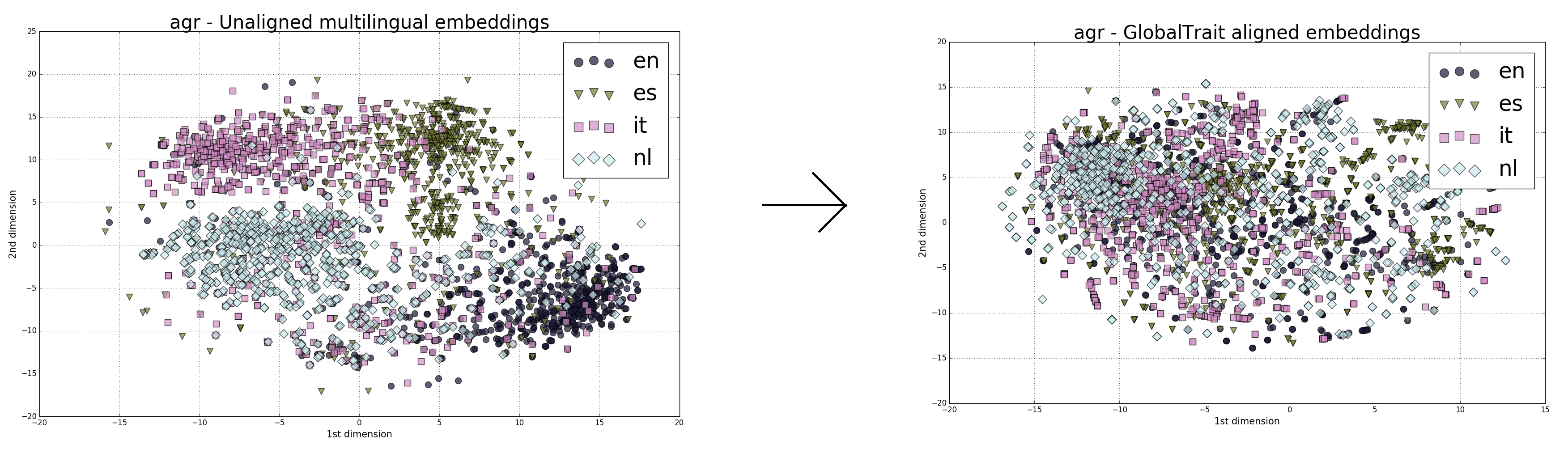}
\caption{Words corresponding to the Agreeableness trait, left shows the unaligned multilingual embeddings, and right shows the GlobalTrait Personality aligned embeddings. We see that the alignment brings the multilingual embeddings to a common vector space (rather than being clustered per language), so we can get the personality trait based mapping between languages.}
\label{agr}
\end{figure*}

\subsubsection{Training Procedure}

We use the same training approach as MUSE to train our personality mapping. For each trait, we take the words having the highest significance in the source languages, and train a second mapping of their corresponding multilingual embeddings to the target language space. We can call such mapping of the Agreeableness trait, for example, as $W_{a}$, and therefore, we try to achieve the following equation, $W_{a}A = B$, where $A$ and $B$ are the trained multilingual embeddings of words in the source and target language respectively, which correlate positively to the Agreeableness trait. Let $X=\{x_1,...,x_n\}$ be the n multilingual embeddings in the source language of the words corresponding most positively to the Agreeableness trait. Likewise, $Y=\{y_1,...,y_m\}$ are the m multilingual embeddings of words in the target language, again for the Agreeableness trait. We try to train a discriminator such that it can differentiate between random samples taken from $W_{a}X = \{W_{a}x_1,...,W_{a}x_n\}$ and $Y$. We train $W_a$ matrix such that the discriminator is unable to differentiate between the two, therefore making the mapping from source as close to the target as possible, bringing the notion of personality based alignment.

\begin{itemize}
    \item \textbf{Discriminator}: Let $\theta_D$ be the parameters of the discriminator, and $P_{\theta_D} (source = 1 | z)$ is the probability that a vector $z$ is the element of a source embedding according to the discriminator, and not the target (mapped) embedding. Therefore, for the Agreeableness trait, the discriminator loss function is as follows:
    \begin{equation}
    \begin{split}
         L_D (\theta_D|W_{a}) = -\frac{1}{n} \sum_{i=1}^n \log P_{\theta_D} (source = 1 | W_{a}x_i) \\ - \frac{1}{m} \sum_{i=1}^m \log P_{\theta_D} (source = 0 | y_i)
    \end{split}
    \end{equation}
    
    \item \textbf{Mapping Matrix}: We try to train $W_a$ such that the discriminator is unable to differentiate between the source and the target (mapped) embeddings:
    \begin{equation}
    \begin{split}
         L_W (W_{a}|\theta_D) = -\frac{1}{n} \sum_{i=1}^n \log P_{\theta_D} (source = 0 | W_{a}x_i) \\ - \frac{1}{m} \sum_{i=1}^m \log P_{\theta_D} (source = 1 | y_i)
    \end{split}
    \end{equation}
    
\end{itemize}

Therefore, the discriminator and the weight matrix objective functions work alternative to each other, and they finally converge in a \textit{min-max} solution. Similarly, we train a mapping matrix for each of the other four traits, which gives us the GlobalTrait personality aligned embeddings, where the embeddings closer in the multilingual vector space reflect the same trait.

\subsection{Convolutional Neural Network}

Deep learning models such as Convolutional Neural networks (CNNs) have gained popularity in the task of text classification \cite{kalchbrenner2014convolutional,kim2014convolutional}. Our CNN model is a two channel mode, where one channel takes the multilingual embeddings, while the other takes the GlobalTrait personality aligned embeddings as input. The first channel with multilingual embeddings is kept trainable, or we can call it a dynamic channel, which means the embeddings are also taken as training parameters, and can change as the training goes on. The other channel, where the personality alignment has been trained already, is kept static. For both channels, we choose window sizes to be 3, 4 and 5, which essentially extracts 3, 4 and 5-gram features from the text, and we have a max pooling operation that keeps the maximum features per window from both the channels. The total features are concatenated and passed to a fully connected layer with a single hidden layer and non-linear activation (tanh), ultimately mapping the features to a binary classification of each trait via softmax.

\section{Experiments}

\begin{table*}[t]
  \caption{F-score results for the monolingual and multilingual performance of the CNN and logistic regression model in the binary classification task for each trait. Bold highlights the best performance per trait for each language.}
  \label{results}
  \centering
  \begin{tabular}{l|c|c|c|c|c|c|c}
  \toprule
       Lang  & Model & Extr & Agr & Cons & Emot & Openn & \textit{Average}              \\
    \hline
  en & Lgr-mono & 64.0 & \textbf{53.3} & 64.5 & 72.5 & 60.5 & 63.0 \\
   & CNN-mono & \textbf{74.4} & 48.2 & \textbf{72.8} & \textbf{74.9} & \textbf{67.7} & \textbf{67.6} \\
   \hline
    & Lgr-mono & 73.2 & 72.1 & 70.0 & 56.9 & 69.0 & 68.2 \\
    & Lgr-multi & 73.2 & 70.1 & 72.6 & 57.0 & \textbf{69.5} & 68.5\\
   es & Lgr-GlobalTrait & 75.9 & 74.7 & 82.4 & 59.0 & 69.0 & 72.2\\
    & CNN-mono & 74.7 & 74.7 & 70.0 & \textbf{69.0} & 67.2 & 71.1\\
    & CNN-GlobalTrait & \textbf{79.4} & \textbf{76.0} & \textbf{83.3} & 67.3 & 67.0 & \textbf{74.6}\\
    \hline
     & Lgr-mono & 53.3 & 60.3 & 52.5 & 60.4 & 63.3 & 58.0\\
       & Lgr-multi & 64.2 & 71.2 & 51.5 & 62.1 & \textbf{64.2} & 62.6\\
     it & Lgr-GlobalTrait & 66.2 & 75.2 & 49.7 & 65.7 & 63.5 & 64.1\\
       & CNN-mono & \textbf{67.2} & 74.3 & \textbf{60.3} & 75.4 & 63.1 & 68.1\\
       & CNN-GlobalTrait & 64.0 & \textbf{77.5} & 58.3 & \textbf{78.0} & 63.2 & \textbf{68.2}\\
    \hline
     & Lgr-mono & 76.0 & \textbf{67.4} & 67.9 & 65.0 & 67.7 & 68.8\\
       & Lgr-multi & 74.2 & 58.2 & 66.8 & 66.2 & 67.0 & 66.5 \\
      nl & Lgr-GlobalTrait & 76.9 & 52.9 & 62.0 & 68.2 & 66.4 & 65.3\\
       & CNN-mono & 76.4 & 60.6 & 61.8 & 78.6 & 64.6 & 68.4\\
       & CNN-GlobalTrait & \textbf{85.3} & 58.4 & \textbf{83.3} & \textbf{85.8} & \textbf{74.5} & \textbf{77.5}\\
       \bottomrule
  \end{tabular}
  
\end{table*}

\subsection{Dataset}

We used the 2015 Author Profiling challenge dataset (PAN 2015) \cite{rangel2015overview}, which includes user tweets in four languages - English (en), Spanish (es), Italian (it) and Dutch (nl), where the personality labels were obtained via self-assessment using the BFI-10 item personality questionnaire \cite{rammstedt2007measuring}. Only the training set was released to us from the PAN2015 website \footnote{\tt https://pan.webis.de/clef15/pan15-web/ \\ author-profiling.html}. Their test data was not available to us as we did not take part in the Author Profiling competition of 2015. The dataset consists of 152 English (14,166 tweets), 110 Spanish (9,879 tweets), 38 Italian (3,687 tweets), and 34 Dutch (3,350 tweets) users in total. Our task is to identify personality in user-level, so we concatenated all the tweets made by a single user to create one single training/test data point. As preprocessing, we tokenized each tweet using Twokenizer \cite{owoputi2013improved}, and replaced all usernames and URL mentions with generic words (\textit{@username} and \textit{@url}), so the model is not affected by mentions in the tweet that are not influenced by the user's personality. Since we are interested in the binary classification of each big five trait, we carried out a median split of the scores, to obtain positive and negative samples (users) for each of the five traits. For our results shown, we carried out a stratified k-fold cross validation, by making k=5 splits of the training set into training/validation, and then show the average result across the 5 different validation sets.

\subsection{Experimental Setup}

For our MUSE training, we used a discriminator with 2 hidden layers, each having a dimension of 2048, and we ran our training for 5 epochs with 100,000 iterations in each epoch. When training the personality alignment, we took the top 3000 significant words corresponding positively to each trait per language. For our evaluation, we built a source to target language dictionary and used mean cosine distance as the validation metric. For our CNN model, we used 64 filters per filter size, and for our fully connected layer, we set the hidden layer dimension to 100, and we ran our training for 100 epochs for each model with batch size = 10. We used binary cross entropy as our loss function, and used Adam optimizer with learning rate of $1e^{-4}$.

\subsection{Text Personality Analysis - Visualization of Embeddings}

To check if words having similar semantic information across languages contribute to similar Big Five traits, we carried out some text-based analysis of our data. For each of the Big Five traits, we first obtained the most significant words in each language using term frequency-inverse document frequency (\textit{tf-idf}) features. We then took the top 750 words for each language having the highest \textit{tf-idf} and plot their trained multilingual embeddings (trained using MUSE from monolingual embeddings in each language) on a 2-D space by performing t-distributed Stochastic Neighbor Embedding (\textit{t-SNE}) on the 300-dimensional vectors. As an example, plots for the Extraversion and Conscientiousness traits are shown in Figure \ref{extr}.

As we can see in the figure, for both traits there is very little overlap in the embedding space between the four languages, and most of the words are clustered per language. This shows that words corresponding to each trait might not have the same semantic meaning across the languages. Similar to the Extraversion and Conscientiousness trait, in all five traits, we see a similar trend, some overlap between English and Spanish, some overlap between Spanish and Italian, but very little or no overlap between Dutch and any of the languages. Therefore, this gives rise to the need for a certain mapping from each language to our target language, English. Some examples of words corresponding positively to the traits per language is shown in table \ref{words}.

\subsection{Binary Classification}

\begin{table*}[t]
    \centering
    \caption{Regression results of our multilingual CNN model using the GlobalTrait aligned embeddings, compared to two of the previous papers' work on the same dataset. Bold indicates best performance (lowest RMSE) per language per trait.}
    \begin{tabular}{l|c|c|c|c|c|c|c}
    \toprule
         Lang & Model & Extr & Agr & Cons & Emot & Openn & \textit{Average} \\
         \hline
          & Char Bi-RNN \cite{liu2016language} & 0.148 & \textbf{0.143} & 0.157 & 0.177 & \textbf{0.136} & 0.152\\
         es & tf-idf linear regression \cite{sulea2015automatic} & 0.152 & 0.148 & \textbf{0.114} & 0.181 & 0.142 & \textbf{0.147} \\
          & \textit{CNN-GlobalTrait} & \textbf{0.142} & 0.150 & 0.135 & \textbf{0.169} & 0.151 & 0.149 \\
          \hline
          & Char Bi-RNN \cite{liu2016language} & 0.124 & 0.130 & \textbf{0.095} & \textbf{0.144} & 0.131 & 0.125 \\
          it & tf-idf linear regression \cite{sulea2015automatic} & 0.119 & \textbf{0.122} & 0.101 & 0.150 & \textbf{0.130} & \textbf{0.124} \\
          & \textit{CNN-GlobalTrait} & \textbf{0.107} & 0.128 & 0.120 & 0.147 & 0.134 & 0.127 \\
          \bottomrule
    \end{tabular}
    \label{regression}
\end{table*}

We first carried out binary classification of the users based on the median split of scores in each trait. Classification is of more importance to us as dialogue systems and other similar applications require us to classify each person into positive/negative for each trait, which can then be used to make decisions such as adapting to the given personality.

\subsubsection{Baseline}

We implemented a simple logistic regression classifier, which takes in the average embeddings of the words as input features, in order to compare our aligned embeddings result with the monolingual counterpart. We present the results of the following experiments for comparison: 

\begin{itemize}
    \item \textbf{Lgr-mono}: logistic regression using monolingual embeddings
    \item \textbf{Lgr-multi}: logistic regression using unaligned multilingual embeddings
    \item \textbf{Lgr-GlobalTrait}: logistic regression using our personality aligned multilingual embeddings
    \item \textbf{CNN-mono}: CNN using monolingual embeddings as input
    \item \textbf{CNN-GlobalTrait}: two channel CNN using multilingual embeddings plus the GlobalTrait aligned embeddings
\end{itemize}

In both `-multi' and `-GlobalTrait', models, the training set includes both the English and the source language's training data, and is tested on the source language's validation sets, while `-mono' is just the monolingual model for the respective source language. Results are reported in table \ref{results}.

\subsubsection{Results}

We achieve an average F-score of 74.6 in Spanish, 68.2 in Italian, and 77.5 in Dutch when using our multilingual CNN model with the GlobalTrait aligned embeddings, which are the highest performance achieved in each of the three languages. As we can see in table \ref{results}, CNN-GlobalTrait performs the best except for two traits in Spanish and three traits in Italian. The discrepancies can be due to the imbalanced nature of the dataset, and the logistic regression being a simple classifier, can converge better than CNN, especially for smaller datasets. For logistic regression, using multilingual and then GlobalTrait aligned embeddings improves on the monolingual results. In general, our multilingual results perform better than monolingual, except for one trait in Spanish, two traits in Italian, and one trait in Dutch. This shows that we can use the features retrieved from English to help us recognize personality in the other languages, and our personality alignment makes it easier for such kind of transfer learning.

\subsection{Regression}

We also carried out regression experiments to compare our model with a recent paper \cite{liu2016language} that tries to perform multilingual personality recognition on the same dataset. They use a character to word to sentence for personality traits (C2W2S4PT) model, which uses a two layer bi-directional RNN model with Gated Recurrent Units (GRU) followed by a fully connected layer to achieve the results.

For our regression task, we used the scores given in the dataset, and did not carry out the median split anymore. We kept our same training procedure as our classification task to train the multilingual and the personality aligned embeddings using MUSE. Our CNN model was also the same except for our last fully connected layer, where we did not have the softmax layer and instead of cross entropy, we used mean-squared error as our objective (loss) function:

\begin{equation}
    L(\theta) = \frac{1}{n} \sum_{i=1}^n (y_{t_i} - \hat{y}_{t_i})^2
\end{equation}

where $y_{t_i}$ is the ground truth personality score of the $t_i$ tweet, and $\hat{y}_{t_i}$ is the predicted score, $\theta$ being the collection of all parameters being trained. As our evaluation metric, we use Root Mean Squared Error (RMSE), which tries to measure the performance via the average error of the model across all users:

\begin{table*}[t]
    \centering
     \caption{Binary classification F-score results on Chinese dataset, showing the comparison of GlobalTrait to other monolingual and multilingual models. All the multilingual models are trained using both Chinese and English training set, and tested on the Chinese validation sets (results shown are average of 5 different train/validation splits).}
    \begin{tabular}{l|c|c|c|c|c|c|c}
    \toprule
         Lang & Model & Extr & Agr & Cons & Emot & Openn & \textit{Average} \\
         \hline
          & Lgr-mono & 58.2 & 59.0 & 57.4 & 56.9 & 56.5 & 57.6 \\
    & Lgr-multi & 62.1 & 60.0 & 61.4 & 60.5 & 58.2 & 60.4 \\
   ch & Lgr-GlobalTrait & 64.1 & \textbf{62.1} & 61.3 & 62.9 & 59.1 & 61.9 \\
    & CNN-mono & 60.6 & 58.4 & 59.3 & 58.2 & 57.5 & 58.8 \\
    & CNN-GlobalTrait & \textbf{64.2} & 61.9 & \textbf{63.0} & \textbf{62.5} & \textbf{60.1} & \textbf{62.3}\\
          \bottomrule
    \end{tabular}
    \label{chinese}
\end{table*}

\begin{equation}
    RMSE_{user} = \sqrt{\frac{1}{n} \sum_{i=1}^U (y_{user_i} - \hat{y}_{user_i})^2}
\end{equation}

where $y_{user_i}$ and $\hat{y}_{user_i}$ are the true and predicted personality trait score of the $i^{th}$ user, and $U$ is the total number of users. Table \ref{regression} shows our results compared to the \cite{liu2016language} model (we only show our results for Spanish and Italian, since they did not use the Dutch data in their paper). We also compare our results to the PAN 2015 participants \cite{sulea2015automatic} who used character n-gram based \textit{tf-idf} features to train a regression model, and achieved one of the highest results in the competition.

As we can see from our results in table \ref{regression}, our model gets the best performance in Extraversion trait for both languages, and it performs comparably in the other traits. This could mean that our GlobalTrait personality alignment works better for Extraversion, such that the words in different languages that correspond positively to Extraversion are indeed closer together in the multilingual vector space. However, it does not perform as well in the Openness trait, for example. It is important to take into consideration that \cite{sulea2015automatic} uses a monolinugual model for each language, and therefore is not expandable to multiple languages. Also, unlike \cite{liu2016language} our model does not use character based RNN, which enables us to train on languages that do not share the same characters as English. To show this, we carried out separate classification experiments on a Chinese dataset.

\subsection{Personality Classification on Chinese}

Personality labeled data is currently rare in languages such as Chinese, which necessitates a model like GlobalTrait, enabling us to use English as additional training data to help us recognize personality in the Chinese test set. We use a Chinese personality labeled dataset called the BIT Speaker Personality Corpus \cite{zhang2017social}, collected and released to us by the Beijing Institute of Technology. It consists of 498 Chinese speech clips, each around 9-13 seconds and labeled with Big Five Personality scores given by five judges. We take a mean of the five scores for each clip, and carry out a median split for our binary classification task on each trait. We use an automatic speech recognition (ASR) system to get the speech transcriptions, and use Jieba segmenter \footnote{\tt https://github.com/fxsjy/jieba} to tokenize the Chinese text into words, since words in Chinese are not separated by a blank space.

We get pre-trained monolingual embeddings of Chinese from fastText and use MUSE to train multilingual embeddings in English and Chinese. We then use our GlobalTrait alignment method to map the positively correlated Chinese words in each trait to our target language space of English. The average of our 5 fold cross-validation experimental results are shown in table \ref{chinese} and we train the same five models that were defined earlier. The results show us that, using our GlobalTrait aligned embeddings undoubtedly improves performance on the Chinese evaluation, which indicates a connection between the English and Chinese data captured via the personality alignment.

\section{Final Discussion and Future Work}

We have seen from our results that using the larger data available in English, we are able to improve our multilingual results, when applied to other languages such as Spanish, Italian, Dutch and Chinese. The personality alignment is particularly interesting, as it shows us how the words used to express different personality traits compare and contrast between multiple languages. Since we train a mapping for each trait per language, one word can have five different embeddings, based on the five different trait mappings. For example, our analysis show that the mapped embedding of the word `mundo' (world) in Spanish is closest to the words `travel, flights, fresh', etc. in English, for the Openness trait, while the same word `mundo' for the Extraversion trait gets closest to `parties, love, life', etc. 

We plan to explore more in the future to get more insight into such mappings for our GlobalTrait alignment, and also apply our model to other datasets in different languages. It will also be interesting to apply our GlobalTrait aligned embeddings to other models such as the Bi-directional RNN model we saw implemented by \cite{liu2016language}, and other hierarchical attention networks, where our sequential data will be the tweets of a single user, based on the chronological order of the user tweets. Another work would be to include a much larger English personality labeled database, possibly the Facebook data released by myPersonality project \footnote{\tt http://mypersonality.org/}, where they collected data of around 154,000 users, with a total of 22 million status updates. Such a large database will help us find a better relation from other languages to English, thereby giving us a more meaningful personality alignment.

\section{Conclusion}

We propose the use of personality aligned embeddings, GlobalTrait, which maps the embeddding space from the source language to our high-resource target language (English), thereby enabling us to get better multilingual results. We have shown in our paper that conventional methods that try to use monolingual or even multilingual word similarity for personality recognition may not always give better results, as words corresponding to personality traits might not have similar semantic meaning across multiple languages. Such a method like GlobalTrait can give us a better understanding of how people express personality across different cultures and languages, and therefore enable us to train better language-independent models for multilingual personality recognition.

\section*{Acknowledgements}

This work was partially funded by CERG \#16214415 of the
Hong Kong Research Grants Council and RDC \#1718050-0 of EMOS.AI.

\bibliography{}

\end{document}